\documentclass[runningheads]{llncs}
\usepackage{graphicx}
\usepackage{subfig}
\usepackage{bbm}

\usepackage{tikz}
\usepackage{comment}
\usepackage{amsmath,amssymb} 
\usepackage{color}
\usepackage{mathtools}
\usepackage{xspace}

\newcommand{\mbp}{\mathbf{p}}

\newcommand{\mbw}{\mathbf{w}}

\newcommand{\mbz}{\mathbf{z}}

\newcommand{\mbI}{\mathbf{I}}

\newcommand{\ignore}[1]{}

\makeatletter
\DeclareRobustCommand\onedot{\futurelet\@let@token\@onedot}
\def\@onedot{\ifx\@let@token.\else.\null\fi\xspace}

\makeatother

\usepackage[accsupp]{axessibility}  

\usepackage[width=122mm,left=12mm,paperwidth=146mm,height=193mm,top=12mm,paperheight=217mm]{geometry}

\begin{document}

\pagestyle{headings}
\mainmatter

\title{IFaceUV: Intuitive Motion Facial Image Generation by Identity Preservation via UV map} 

\titlerunning{IFaceUV: Intuitive Motion Facial Image Generation by Identity Preservation via UV map}
%
\author{Hansol Lee \and
Yunhoe Ku  \and
Eunseo Kim \and
Seungryul Baek}
\authorrunning{Lee et al.}
%
\institute{Ulsan National Institute of Science and Technology (UNIST), South Korea
\email{\{hansollee, gyh0316, eunseo and srbaek\}@unist.ac.kr}}

\maketitle 
\begin{abstract}
Reenacting facial images is an important task that can find numerous applications. We proposed IFaceUV, a fully differentiable pipeline that properly combines 2D and 3D information to conduct the facial reenactment task. The three-dimensional morphable face models (3DMMs) and corresponding UV maps are utilized to intuitively control facial motions and textures, respectively. Two-dimensional techniques based on 2D image warping is further required to compensate for missing components of the 3DMMs such as backgrounds, ear, hair and etc. In our pipeline, we first extract 3DMM parameters and corresponding UV maps from source and target images. Then, initial UV maps are refined by the UV map refinement network and it is rendered to the image with the motion manipulated 3DMM parameters. In parallel, we warp the source image according to the 2D flow field obtained from the 2D warping network. Rendered and warped images are combined in the final editing network to generate the final reenactment image.
Additionally, we tested our model for the audio-driven facial reenactment task. Extensive qualitative and quantitative experiments illustrate the remarkable performance of our method compared to other state-of-the-art methods.
\end{abstract}

\section{Introduction}
Face reenactment is an facial image/video synthesizing task that combines the appearance features extracted from the still source images and the motion patterns extracted from a driving video~\cite{NEURIPS2019_31c0b36a}. For example, a face of a certain person in an image can be animated by following the facial appearance of another person as in Fig.~\ref{fig1}. Developing the facial reenactment method in a controllable way has numerous applications including interactive systems, photography, movie production, image enhancement/editing, gaming, telepresence, AR/VR applications, and so on. However, generating the photo-realistic facial images is a challenging task, due to the complicated photo-metric and geometric features such as mouth cavity, hair, and surrounding clothing. 
This is practically more challenging since it requires the model to recognize reliable 3D geometric representations of given human faces. 

Most previous studies in the literature try to solve the problem with computer graphic methods using texture maps and mesh-based surface techniques~\cite{thies2016face2face}. Recently, advances in the deep generative algorithms have emerged as an effective tool for the task~\cite{wiles2018x2face,balakrishnan2018synthesizing,bansal2018recycle,zablotskaia2019dwnet,zakharov2019few}. Particularly, variational auto-encoders (VAEs)~\cite{kingma2013auto} and generative adversarial networks (GANs)~\cite{goodfellow2014generative} have extensively been used in synthesizing naturalistic faces~\cite{brock2018large,choi2018stargan,karras2019style} and to transfer head motion patterns~\cite{bansal2018recycle} or facial expressions~\cite{wang2018video} allying various subjects in the videos.

Recently, neural network-based methods have been proposed to synthesize facial images~\cite{suwajanakorn2017synthesizing,kim2018deep,lombardi2018deep}. Some recent studies~\cite{wang2019few,ha2020marionette,tripathy2020icface,fu2021high} proposed methods that are able to generate talking heads from a single image or multiple images. 
Some of the previous methods got better generalization by inserting neural textures of sources to the targets \cite{wang2019few,burkov2020neural} or defining effective deformation modules~\cite{zakharov2019few,chen2020puppeteergan,NEURIPS2019_31c0b36a}. Although, most previous studies applied subject-specific and indirect motion descriptors like semantic segmentation~\cite{chen2020puppeteergan}, edges~\cite{wang2019few,ren2020deep}, or keypoints~\cite{NEURIPS2019_31c0b36a} to define the target motion.

In this work, we propose IFaceUV, a model for intuitive motion facial image generation. 
We use intuitive 3DMM parameters as target motions to synthesize realistic face images. We used 2D and 3D information appropriately to create realistic images while delivering accurate motion. In order to take full advantage of 3D information, we used the UV maps in the reenactment task to get the accurate motion of the target image and preserve the identity of the source image. And in order to properly change the image corresponding to parts other than the face, a background image was created using a 2D warping network similar to the previous paper~\cite{ren2021pirenderer}. We employ four modules end-to-end to generate facial images. The Data preprocessing module extracts 3DMM parameters from the target image and creates a motion latent vector to inject into the networks. The 2D warping module transforms the 2D image appropriately with the source image and the extracted motion latent vector. The foreground face generating module makes a deformed facial image according to the target motion. It uses the 3DMM information to acquire the 3D face mesh of the target motion and the UV map refinement network to create a corresponding realistic UV map and then generates an image of the face part. Finally, the images obtained above are put into the final editing module to get the final image. And additionally, we performed an audio-driven reenactment task. We use a model that extracts 3DMM parameters from an audio input source and confirm the scalability by conducting an experiment to reenact with only the source image and audio. The main contributions can be summarized as follows:
\begin{itemize}
\item We propose the end-to-end facial reenactment pipeline that combines the 2D and 3D information. We use the 3D information from the 3DMM model and the 2D information from the image warping techniques to compensate for each other.
\item We propose to improve the quality of the UV map and the effective combining methods for the rendered and warped 2D images, thereby generating the realistic final images.
\item By the extensive comparison to several state-of-the-art facial reenactment methods, we demonstrated the superiority of our algorithm.
\end{itemize}

\section{Related Work}
\subsection{3D Morphable Face Models}
Blanz and Vetter~\cite{blanz1999morphable} introduced linear 3D morphable models (3DMM), which solve dense correspondence problem by converting the 3D surface into 2D space. Since then, Multiple variants of 3DMM~\cite{booth20163d,paysan20093d,vlasic2006face,booth20173d,koppen2018gaussian} have been introduced. Generally, 3DMM creation consists of two phases such as; the construction of group by group dense correspondence of each facial mesh in a training set and then applying some statistical analysis on the trained data to get a 2D model.

Booth et al.~\cite{booth20163d} used 10,000 facial scans to obtain a more detailed PCA model. Paysan et al.~\cite{paysan20093d} enhanced the performance of previous models by employing superior scanning equipment and using nonrigid iterative closest point for registration instead of UV space alignment. Vlasic et al.~\cite{vlasic2006face} employed a multi-linear 3D face mesh model that parameterizes the space of geometric variations owing to various characteristics individually. Booth et al.~\cite{booth20173d} employed a more advanced model that included textural differences found in nature. 
To describe the global population as a mixture of Gaussian subspaces, Koppen et al.~\cite{booth20173d} introduced 3DMM and Gaussian mixture.

The Basel Face Model (BFM)~\cite{paysan20093d} is used to recover 3D facial shapes and coarse textures. Specifically, it uses the optimal step nonrigid iterative closest point technique to register a template mesh to the scanned faces, then uses PCA for dimensionality reduction to build the model. We will use this model as a 3DMM model to get the 3D face information we need.

\subsection{3D facial texture}
Deep learning-based approaches have recently been used to reconstruct facial textures from single image. Deng et al.~\cite{deng2019arcface}, for example, proposed a method for concurrently predicting the 3DMM shape and texture coefficients that employs an illumination and rendering model during training and performs image-level and perception-level losses, yielding a better result than others. However, the generated textures are still limited because they are generated by the 3DMM explicit model.
These approaches~\cite{gecer2019ganfit,deng2018uv,gecer2020synthesizing} offer the possibility to train face textures in UV space using Generative Adversarial Networks (GANs) to refine the textures of the 3D Face Morphable Model (3DMM).

On an unwrapped UV space, Gecer et al.~\cite{gecer2019ganfit} demonstrated a high-resolution statistical reconstruction of face textures. However, it's hard to compile a large database of high-resolution UV maps in the wilds scenarios. As a result, they are unable to adapt to the wild. To handle this issue, Deng et al.~\cite{deng2018uv} represented an approach for completing the UV map extracted from the in-the-wild images. They collected complete UV map by fitting 3DMM on various multi-view images and video data sets. And then, they proposed an architecture that combines local and global adversarial DCNNs to develop a facial UV completion model that preserves identity.
 
Gecer et al.~\cite{gecer2020synthesizing} proposed a method to handle the problem of not being able to faithfully express the texture of the face or the norm of the face. They suggested an approach for jointly generating high-quality textures, shapes, and norms that can be used for photo-realistic generation. To this end, they proposed a novel GAN that can generate data from different forms while using correlations. In this paper,  we generate realistic UV maps that are suitable for reenactment by injecting motion vectors along with images into the network. 

\subsection{Facial Reenactment}
The face reenactment aims to translate a source image to another pose-and-expression, which is provided by the driving image.
Recent works demonstrate great success and advance in face reenactment. For instance, 
\textit{Face2Face}~\cite{thies2016face2face} animates the facial expression of the source video by utilizing the rendered image.
\textit{ReenactGAN}~\cite{wu2018reenactgan} drives a specific identity with the encoder-decoder framework. 
Similarly, \textit{X2face}~\cite{wiles2018x2face} uses two networks, embedding network which is U-Net based structure to represent embedded face correctly, and driving network which is also encoder-decoder structure to well represent latent vector which encodes pose, expression, and so on. 

The work of~\cite{kim2018deep} transfers the full 3D head position, head rotation, face expression, eye gaze, and eye blinking from a driving actor to a portrait video of the source actor. 
\textit{FSGAN}~\cite{nirkin2019fsgan} also performs subject agnostic face reenactment which adjusts for both pose and expression variations. 
\textit{Bi-layer}~\cite{zakharov2020fast} can make realistic results by using a coarse-to-fine manner, specifically, it adds warped pose-independent high-frequency texture to pose-dependent low-frequency component so that it can generate the flawless results. 
\textit{HeadGAN}~\cite{doukas2021headgan} takes data preprocessing, which uses knowledge of 3DMM~\cite{blanz1999morphable} to make 3D face rendering, to precisely transfer expression of driving face and identity of source face. And with additional audio information, it can make photo-realistic results in a one-shot face reenactment method. 

For multi identity face reenactment, Huang~\cite{huang2020learning} proposed \textit{CrossID-GAN}, which extracts and transfers ID-invariant landmark information across face images of people with different identities. 
Also, \textit{FReeNet}~\cite{zhang2020freenet} introduces Unified Landmark Converter (ULC) and Geometry-aware Generator (GAG) which represent facial expression to landmark space. In recent, \textit{PIRenderer}~\cite{ren2021pirenderer} progresses facial reenactment using warping network, refer to 3DMM~\cite{blanz1999morphable} parameters. The warping network is performed in two-dimensional, it can generate realistic images. However, a two-dimensional network is weak about making hidden areas of the face. In our method, we used a realistic UV map to robust about creating hidden areas.

\subsection{Audio-driven Reenactment}

The goal of audio-driven reenactment is to generate an image-realistic video stream using a source image and an input audio stream. There are numerous methods for this task but most of them directly apply the relationship between the source image and audio stream~\cite{chung2017you,vougioukas2018end,zhou2019talking,vougioukas2020realistic}. Chung et al.~\cite{chung2017you} proposed a method to animate the face of a normalized and non-moving image to follow an input audio stream. They first projected the source image and target audio into a latent space with a deep encoder and then apply combined embedding of both face and audio using the decoder to incorporate a talking head. Vougioukas et al.~\cite{vougioukas2018end} presented a 2D image-based technique that uses a temporal GAN to generate a talking face video from a source image and audio stream as input.

Zhou et al.~\cite{zhou2019talking} proposed a method that integrates image and audio features by learning the disentangled audio-visual representation through an associative-and-adversarial process. Vougioukas et al.~\cite{vougioukas2020realistic} represented a temporal GAN network that uses three discriminators focused on getting detailed input frames, audio-visual synchronization, and logical expression. Since all these methods use direct modeling of the image and audio features so they cannot learn the head poses of the source image, thus generating output videos with a fixed head pose.

To avoid this issue, only a few methods are presented that can generate natural, photo-realistic, and full-frame videos~\cite{suwajanakorn2017synthesizing,thies2020neural,ren2021pirenderer}. Suwajanakorn et al.~\cite{suwajanakorn2017synthesizing} generated a high-quality video of President Barack Obama using an audio stream of him. An RNN is trained to learn the sequence of the mouth shape from many hours of his audio speech. Thies et al.~\cite{thies2020neural} proposed a temporal DNN architecture to map an input audio signal to a 3D blendshape that can generate a video with a specific talking style. Ren et al.~\cite{ren2021pirenderer} proposed a method that extracts a sequential motion of a face from the input audio stream and generates a video with convincing head rotation from a single source image.

\section{Method}
\begin{figure*}[t!]
\centering
\includegraphics[width=1\linewidth]{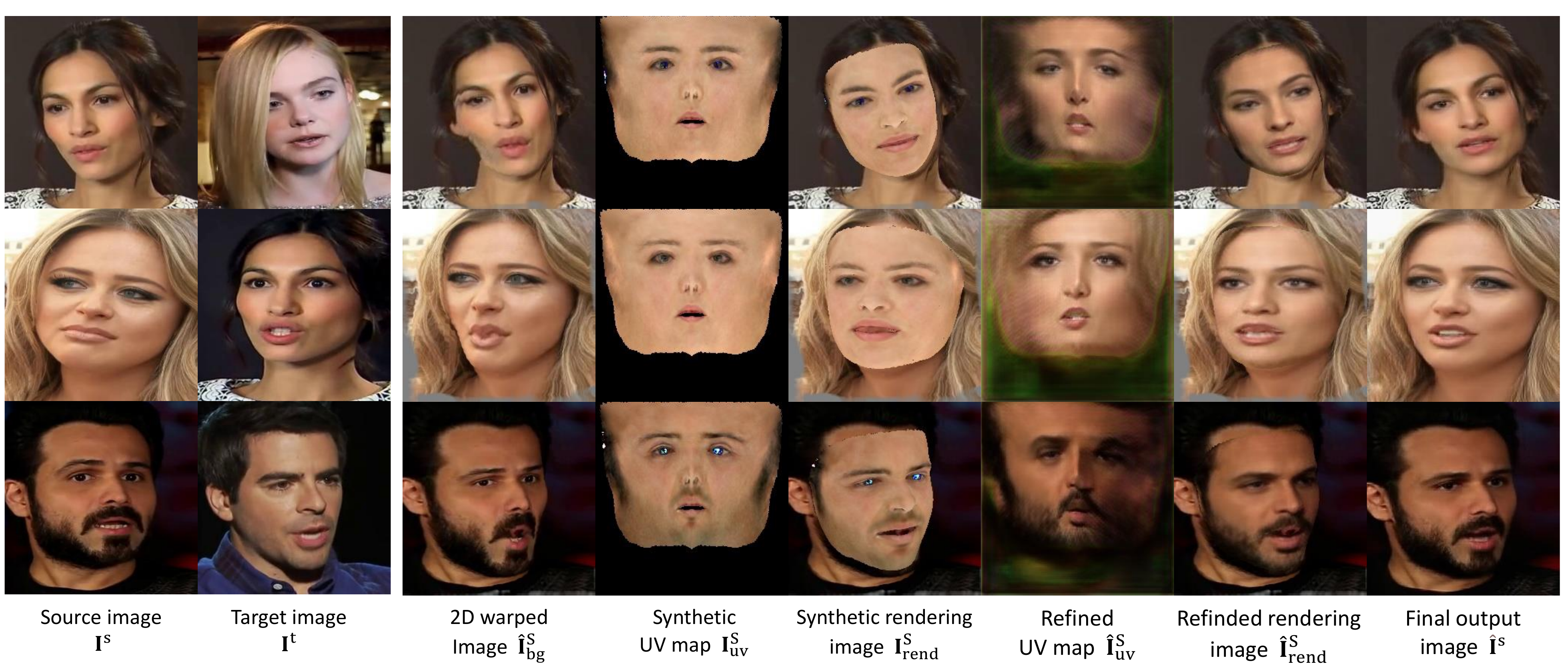}
\caption{\textbf{Examples of images generated from our method.} The source images are in the first column, and the target images, from which poses and expressions are to be derived, are in the second column. The process of creating the final output images shown at the end is described in columns three through seven. The 2D warped images based on the target motion are in the third column. The UV maps generated by~\cite{deng2019accurate} are the fourth. The refined UV maps are in the next column. The following column shows the combined images of the refined UV maps and the generated 2D warped images.}

\label{fig1}
\end{figure*}

\begin{figure*}
\centering
\includegraphics[width=1\linewidth]{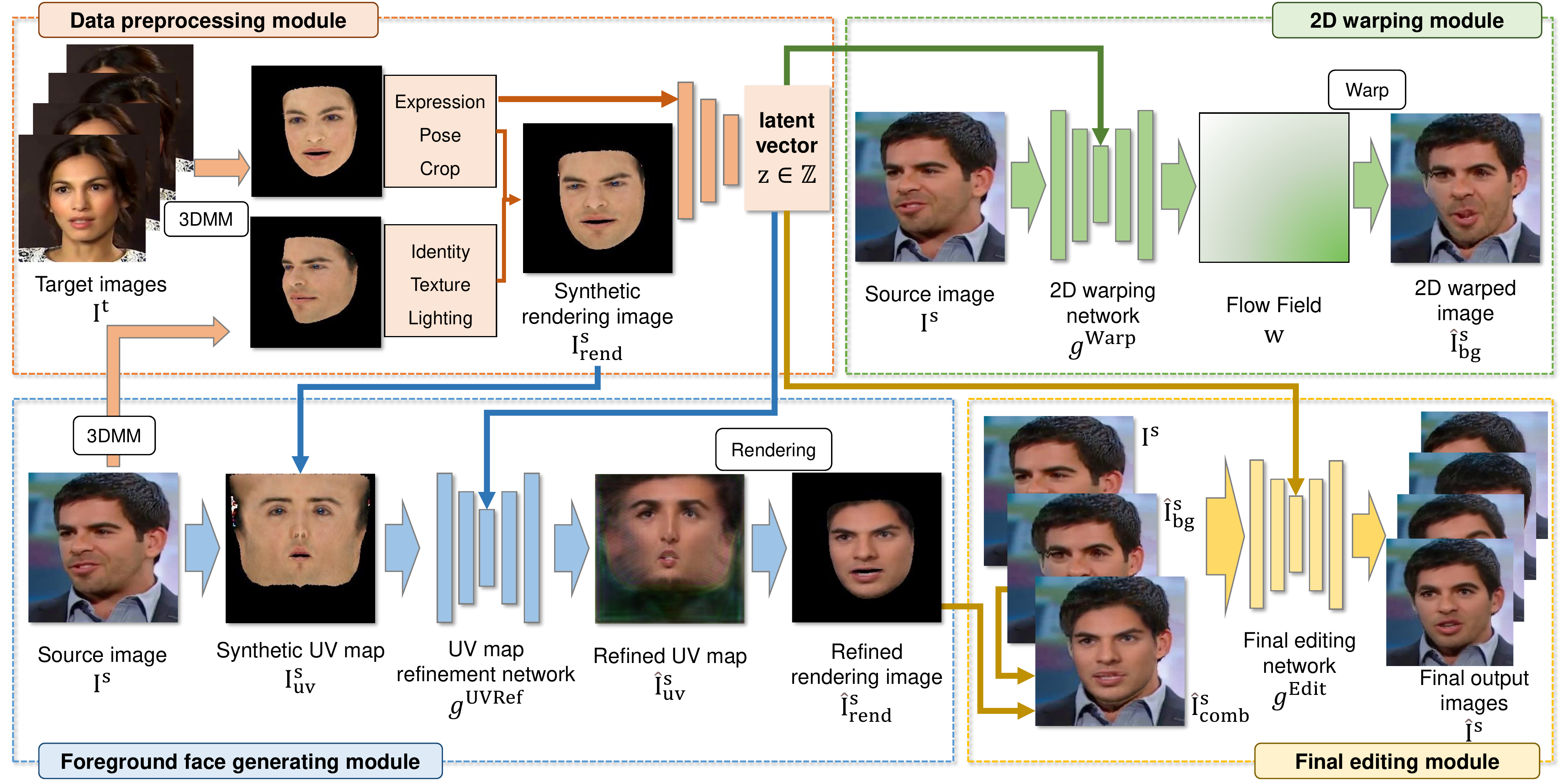}
  \caption{\textbf{The overall framework of our method.} First, the target motion latent vector $\mbz$ is created from the data preprocessing module from the target images. And we put this latent vector as input motion to each module. In the 2D warping module, the flow field and the corresponding image are generated from the source image $\mbI^s$ and the target motion latent vector. And UV map refinement network generate refined UV map $\hat{\mbI}_\text{uv}^\text{s}$. In the end, the final editing network create final output images $\hat{\mbI}^\text{s}$.}
  \vspace{-0.3cm}
\label{framework}
\end{figure*}

In this paper, we denote the images as $\mbI^{\text{x}}_{\text{y}} \in\mathbb{R}^{3\times256\times256}$, where $\text{x}$ denotes a specific person in an image (i.e., $\text{s}$ is the source person and $\text{t}$ is the target person) and $\text{y}$ denotes the type of image (for instance, $\mbI^{\text{s}}_{\text{uv}}$ is UV map image with source person). By using 3D facial information, we are able to properly reenact the 2D face image by leveraging accurate facial expressions and poses of the target image $\mbI^{\text{t}}$ to the source image $\mbI^{\text{s}}$ while keeping the identity of the source image. Additionally, we warp images using 2D information in order to create a realistic 2D image $\hat{\mbI}^{\text{s}}$. Utilizing a 3D face model and a related UV map, the face component was generated in better detail while retaining its identity. Before we introduce the specifics of our strategy, Fig.~\ref{fig1}. outlines what we were attempting to accomplish. And our overall framework is described in Fig.~\ref{framework}.

\subsection{Data preprocessing module}
Data preprocessing module extracts 3DMM parameters~\cite{paysan20093d} from both source and target images $\mbI^\text{s}$ and $\mbI^\text{t}$, respectively. Then, we replace the source image $\mbI^\text{s}$'s motion parameter with the target image $\mbI^\text{t}$'s. Finally, we apply the encoder $f^\text{E}$ to map it towards the latent vector $\mbz\in Z\subset\mathbb{R}^{1\times 256}$.

The 3DMM motion parameters are composed of three distinct vectors: expression $\mbp^\text{exp}\in\mathbb{R}^{1\times 64}$, angle $\mbp^\text{angle}\in\mathbb{R}^{1\times 3}$ and translation $\mbp^\text{trans}\in\mathbb{R}^{1\times 3}$. Among this, we define the motion descriptor as $\mbp_t$ for the $t$-th frame of the video, by concatenating expression and pose vectors as:
\begin{eqnarray}
   \mbp_t = (\mbp_t^\text{exp}, \mbp_t^\text{angle}, \mbp_t^\text{trans}).
\end{eqnarray}
Also, this is extracted from~\cite{deng2019accurate}.

For the simplicity, we will represent angle parameter $\mbp_\text{t}^\text{angle}$ and translation parameter $\mbp_\text{t}^\text{trans}$ as pose parameters $\mbp^{\text{pose}}\in\mathbb{R}^{6}$. Although the 3DMM parameters are reconstructed rather accurately, there is a possibility of the estimation error. In order to reduce the impact of the noise, we extract the 3DMM coefficients from multiple frames as follows:
\begin{eqnarray}
   \mbp \equiv \mbp_\text{t-i:t+i} \equiv \{\mbp_\text{t-i}^\text{exp},\mbp_\text{t-i}^\text{pose}, \ldots,\mbp_\text{t}^\text{exp},\mbp_\text{t}^\text{pose},
\ldots,\mbp_{\text{t+i}}^\text{exp},\mbp_{\text{t+i}}^\text{pose} \}
\end{eqnarray}
by generating extracted 3DMM parameters $\mbp$ into a latent vector $\mbz$. 
To transform 3DMM coefficients to the latent vector $\mbz$, we use an encoder architecture $f^\text{E}:\mbp \rightarrow \mbz$. The obtained latent vector $\mbz$ is injected into the generating networks as the final operation descriptor after passing through the AdaIN procedure. The AdaIN operation is defined as:
\begin{eqnarray}
   \text{AdaIN}(\mathbf {I}_i, \mathbf {z}) = \sigma(\textbf{z}_s) \left (\frac {{\mbI}_i-\mu ({\mbI}_i)}{\sigma (\mathbf {I}_i)} \right) + \mu(\textbf{z}_b)
\end{eqnarray}
The mean and square root of variance opetions are represented by $\mu(\cdot)$ and $\sigma(\cdot)$, respectively. Each feature map $\textbf{I}_i$ is firstly normalized, then scaled and biased by $\sigma(\textbf{z}_s)$ and $\mu(\textbf{z}_b)$, which are respectively determined by fully connected layer output with $\mathbf{z}$ as input. 

\subsection{2D image warping module} \label{2d_warping}

The 2D image warping module $f^\text{Warp}: [\mbI^{\text{s}}, \mbz] \rightarrow \mbw$ is responsible for spatially transforming a source image $\mbI^\text{s}$ into a target image $\mbI^\text{t}$ based on detecting the shape of the source face and estimating the deformations between source and target images called as the warp flow field $\mbw \in \mathbb{R}^{64 \times 64\times 2}$. If angles between two faces in source and target images are significantly different, generated face images and the background image could be distorted. Therefore, we warp the image pixels while changing certain portions of the image like hair and background appropriately.

The 2D warping network $f^\text{Warp}$ has an auto-encoder structure. 
Furthermore, we did not estimate the complete resolution of the flow field $\mbw$ following~\cite{jonschkowski2020matters,teed2020raft,ren2020deep,NEURIPS2019_31c0b36a,ren2021pirenderer}: The output flow has a resolution of $1/4$ compared to that of the input image $\mbI$. During the training and evaluation steps, we upsample the flow field $\mbw$ 4 times to fit to the image resolution. Using the flow field $\mbw$, the warped background image $\hat{\mbI}_{\text{bg}}^{\text{s}} = \mbw(\mbI^\text{s})$ is generated. Since there is no ground-truth flow fields to supervise the 2D warping network $f^\text{Warp}$, we trained the 2D warping network by calculating the loss with the target image, using the perceptual loss~\cite{Johnson2016Perceptual} as follows:
\begin{eqnarray}
   \mathcal{L}_{\text{Warp}}(f^\text{Warp}) = \sum_{i}\Big\| \phi_i(\mbI^\text{t}) -  \phi_i(\hat{\mbI}^\text{s}_{\text{bg}})\Big\|_1
\end{eqnarray}
where $\phi_i$ denotes the $i$-th layer of the VGG network response. The 2D warping network is trained in the  self-supervised way by reducing the difference between 2D warped images $\mbI^\text{s}_{\text{bg}}$ and their corresponding source images $\mbI^\text{s}$.

\subsection{Foreground face generating module}

This module is responsible for generating rendered images $\hat{\mbI}^\text{s}_\text{rend}$ that follow source images' $\mbI^\text{s}$ identity while following target images' $\mbI^\text{t}$ motions. If we na\"ively render 3D meshes obtained from the 3DMM model~\cite{deng2019accurate} into 2D images, as in the $5$th column of the Fig.~\ref{fig1}, the images look quite artificial. In order to generate flawless and realistic facial images, we map the textures of the recovered 3DMM model into the UV map, implement an UV map refinement network
\begin{eqnarray}
    f^\text{UVRef}: [\mbI^\text{s}, \mbI_\text{uv}^\text{s}, \mbz] \rightarrow \hat{\mbI}_\text{uv}^\text{s}
\end{eqnarray}
which generates the realistic UV map $\hat{\mbI}_\text{uv}^\text{s}$ exploiting source image $\mbI^\text{s}$, synthetic UV map $\mbI_\text{uv}^\text{s}$ and latent vector $\mbz$ as inputs.

After obtaining the refined UV map $\hat{\mbI}^\text{s}_\text{uv}$, we adopt the PyTorch3D~\cite{ravi2020accelerating} differentiable renderer $\mathcal{R}$ to render a realistic face image $\hat{\mbI}^{\text{s}}_{\text{rend}}$ as follows:
\begin{eqnarray}
   \hat{\mbI}_\text{rend}^\text{s} = \mathcal{R}(\textbf{M}, \hat{\mbI}_\text{uv}^\text{s})
\end{eqnarray}
where the renderer $\mathcal{R}$ takes the mesh $\textbf{M}$ and the realistic UV map $\hat{\mbI}^\text{s}_\text{uv}$ as inputs and outputs the realistic rendered image $\hat{\mbI}^\text{s}_\text{rend}$.
Then we overlay the rendered realistic image $\hat{\mbI}^{\text{s}}_{\text{rend}}$ to the warped background image $\hat{\mbI}^\text{s}_{\text{bg}}$ to make combined images $\hat{\mbI}^\text{s}_\text{comb}$. The overlay is operated using the facial binary mask $\mbI_\text{m}^\text{t}$ representing the missing pixels on the ground-truth target image $\mbI^\text{t}$ with $\mbI^\text{t}_\text{m}=0$ and the valid region on the ground-truth target image with $\mbI^\text{t}_\text{m}=1$. In conclusion, the combined image $\hat{\mbI}^\text{s}_\text{comb}$ is defined like this:
\begin{eqnarray}
   \hat{\mbI}_\text{comb}^\text{s} = ((1 - \mbI_\text{m}^\text{t} ) \cdot \hat{\mbI}_\text{bg}^\text{s}) \odot (\mbI_\text{m}^\text{t} \cdot \hat{\mbI}_\text{rend}^\text{s})
\end{eqnarray}

Since we could not get a ground-truth image of the UV map, we trained to reduce the differences between the ground-truth target image $\mbI^\text{t}$ and the combination of two images $\hat{\mbI}_\text{comb}^\text{s}$, rendered image and warped image.
We trained our UV map refinement network using the L2 loss $\mathcal{L}_{\text{UVRef}}$:
\begin{eqnarray}
   \mathcal{L}_{\text{UVRef}}(f^\text{UVRef}) = \sum_{i}\Big\| \phi_i(\mbI^\text{t}) -  \phi_i(\hat{\mbI}_\text{comb}^\text{s} )\Big\|_1
\end{eqnarray}
The combined image $\mbI^\text{s}_{\text{comb}}$ generated from that source frame and the  $\mbI^\text{t}$ are the same source and target frames in the 2D warping module, respectively.

\subsubsection{Cross consistency of same person UV map}
We trained the source image as the motion of the target image, on the contrary, we also trained the target image once as the motion of the source image.
Because we used same person as source and target, the same UV map came out.
And because of this reason, we suggested a consistency loss $\mathcal{L}_\text{cons}$ for ensuring the identicality of source and target UV maps:
\begin{eqnarray}
   \mathcal{L}_{\text{cons}}(f^\text{UVRef}) = \sum_{i}\Big\| \hat{\mbI}_{\text{uv}}^\text{s} -  \frac{\hat{\mbI}_{\text{uv}}^\text{s} +  \hat{\mbI}_{\text{uv}}^\text{t}}{2} \Big\|_{1}
\end{eqnarray}
In the equation, $\hat{\mbI}_{\text{uv}}^\text{s}$ is the UV map of source image, and $\hat{\mbI}_{\text{uv}}^\text{t}$ is the UV map of target image. We use the L1 reconstruction loss to make the consistency between these two UV maps.

\subsection{Final editing module} \label{final_editing}

The combined image $\hat{\mbI}_\text{comb}^\text{s}$ is not in the good quality. To improve the quality of the combined image $\hat{\mbI}_\text{comb}^\text{s}$, we involve a final editing network as:
\begin{eqnarray}
f^\text{Edit}: [\mbI^\text{s}, \hat{\mbI}_\text{bg}^\text{s}, \hat{\mbI}_\text{comb}^\text{s}, \mbz] \rightarrow \hat{\mbI}^\text{s}.
\end{eqnarray}
which receives three images $\mbI^\text{s}, \hat{\mbI}_\text{bg}^\text{s}, \hat{\mbI}_\text{comb}^\text{s}$ and the latent vector $\mbz$ as input, and generates the final output image $\hat{\mbI}^\text{s}$. The final editing network has an similar architecture with the UV map refinement network. 
For training the final editing network, we involve the loss function $\mathcal{L}_\text{Edit}$
\begin{eqnarray}
   \mathcal{L}_{\text{Edit}}(f^\text{Edit}) = \sum_{i}\Big\| \phi_i(\mbI^\text{t}) -  \phi_i(\hat{\mbI}^\text{s})\Big\|_1
\end{eqnarray}
where $\phi_i$ denotes the $i$-th layer's response of the VGG network.

This final editing network was also trained through the perceptual loss. We trained this network by reducing the differences between the final output image of this network  $\hat{\mbI}^\text{s}$, which takes the output images of the above networks as input and the target frame image $\mbI^\text{t}$. Total loss of our model is a sum of all losses introduced above, therefore the total loss is defined as: 
\begin{eqnarray}
   \mathcal{L} &=&  \lambda_{\text{Warp}}\mathcal{L}_{\text{Warp}}(f^\text{Warp}) + \lambda_{\text{UVRef}}\mathcal{L}_{\text{UVRef}}(f^\text{UVRef})\nonumber\\ &+& \lambda_{\text{cons}}\mathcal{L}_{\text{cons}}(f^\text{UVRef}) +\lambda_{\text{Edit}}\mathcal{L}_{\text{Edit}}(f^\text{Edit})
\end{eqnarray}
where $\lambda_{\text{warp}} = 2.5, \lambda_{\text{uvref}} = 4, \lambda_{\text{cons}} = 1,$ and $\lambda_{\text{edit}} = 4$.

\subsection{Extension of face reenactment with audio source}
\begin{figure*}[t!]
\centering
\includegraphics[width=1\linewidth]{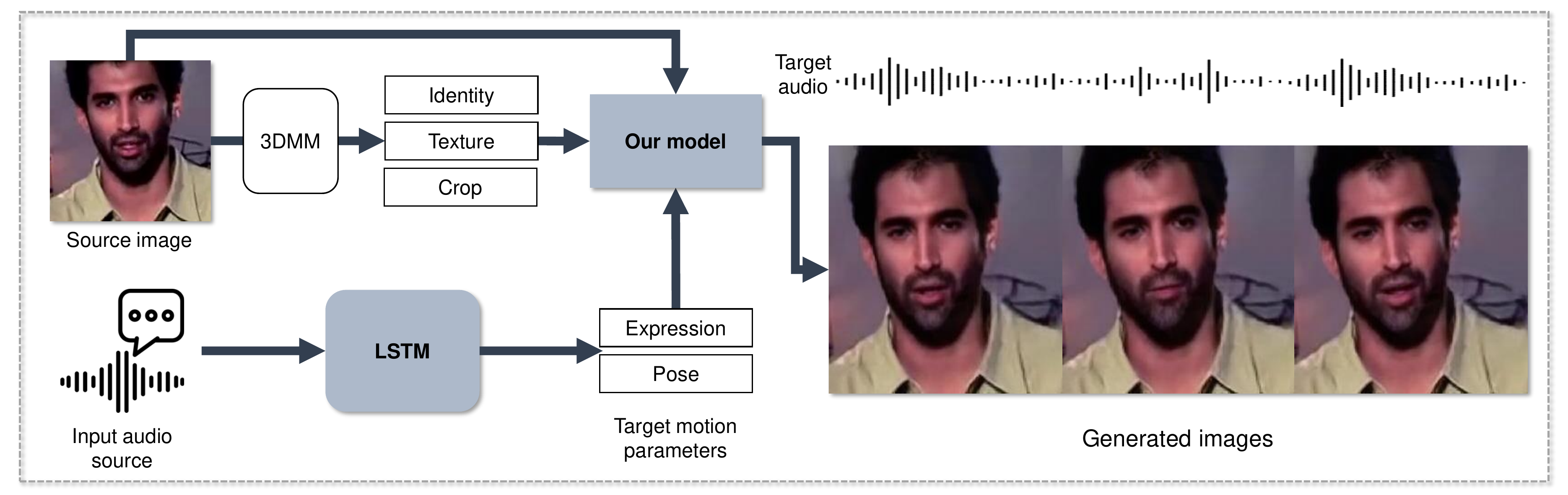}
  \caption{\textbf{The flow of extension of face reenactment with audio source.} First, the audio source passes through the LSTM model to extract expression and pose parameters as target motion. And we make generated images using the source image and target motions as input.}
  \vspace{-0.3cm}
\label{fig:fig3}
\end{figure*}

We can generate a facial image of target motion from the source image as long as there are only the target 3DMM parameters.
For that reason, we can expand the face reenactment task with only a source image and audio input sources.
The process of generating images from the audio source can be checked through Fig.~\ref{fig:fig3}.
We receive 3DMM expression and pose parameters as target motion.
We used an LSTM network to extract 3DMM expression and pose parameters from the audio source.
In order to extract 3DMM expression and pose parameters from audio sources, we used the audio-to-3DMM expression and pose parameters mapping method introduced in this~\cite{yi2020audio}.  We first extracted \textit{Mel-Frequency Cepstral Coefficient} (MFCC) features from the input audio source. Then, the LSTM network was used to extract 3DMM parameters from MFCC features. The LSTM network we used was trained to generate a ground-truth expression parameter sequence $\mbp^\text{exp} = \{\mbp^\text{exp}_{(1)}, ...,\mbp^\text{exp}_{(\text{t})}\}$ and pose parameter sequence $\mbp^\text{pose} = \{ \mbp^\text{pose}_{(1)}, ...,\mbp^\text{pose}_{(t)}\}$ from the MFCC feature sequence $\textbf{f} =  \{ f_{(1)},..., f_{(\text{t})}\} $. And this model trained using with four losses. 
These are two MSE losses and two inter-frame continuity losses.
Two MSE losses are conducted between the generated and ground-truth parameters for expression and pose. And the inter-frame continuity loss is calculated using the squared $L2$ norm of the gradient of each component, expression and pose.

\begin{figure*}[t!]
\captionsetup[subfigure]{labelformat=empty}
\centering
\subfloat[]{\includegraphics[height=0.14\linewidth, width=0.14\linewidth]{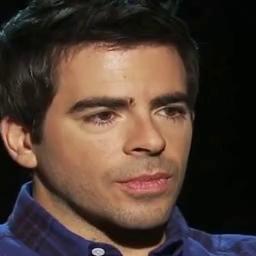}}
\subfloat[]{\includegraphics[height=0.14\linewidth, width=0.14\linewidth]{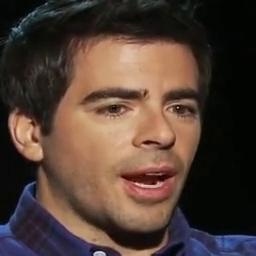}}\hspace{0cm}
\subfloat[]{\includegraphics[height=0.14\linewidth, width=0.14\linewidth]{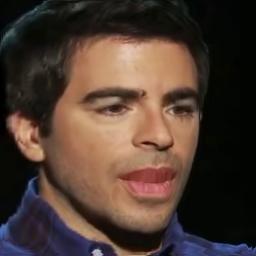}}
\subfloat[]{\includegraphics[height=0.14\linewidth, width=0.14\linewidth]{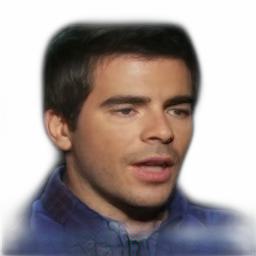}}
\subfloat[]{\includegraphics[height=0.14\linewidth, width=0.14\linewidth]{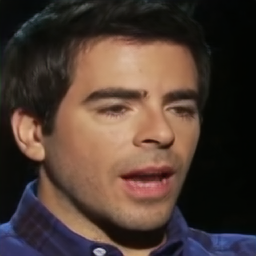}}
\subfloat[]{\includegraphics[height=0.14\linewidth, width=0.14\linewidth]{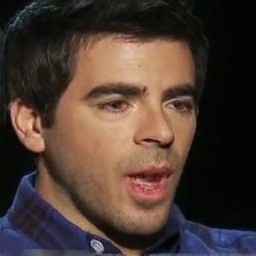}}
\subfloat[]{\includegraphics[height=0.14\linewidth, width=0.14\linewidth]{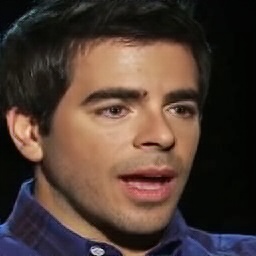}}
\\
\vspace{-0.95cm}
\subfloat[]{\includegraphics[height=0.14\linewidth, width=0.14\linewidth]{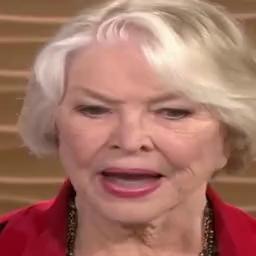}}
\subfloat[]{\includegraphics[height=0.14\linewidth, width=0.14\linewidth]{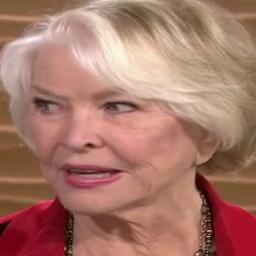}}\hspace{0cm}
\subfloat[]{\includegraphics[height=0.14\linewidth, width=0.14\linewidth]{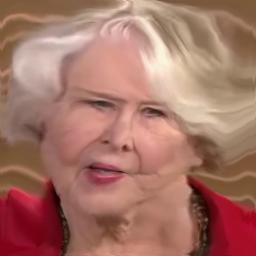}}
\subfloat[]{\includegraphics[height=0.14\linewidth, width=0.14\linewidth]{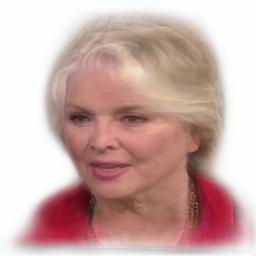}}
\subfloat[]{\includegraphics[height=0.14\linewidth, width=0.14\linewidth]{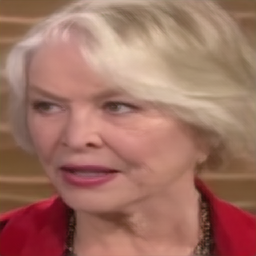}}
\subfloat[]{\includegraphics[height=0.14\linewidth, width=0.14\linewidth]{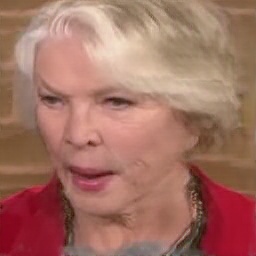}}
\subfloat[]{\includegraphics[height=0.14\linewidth, width=0.14\linewidth]{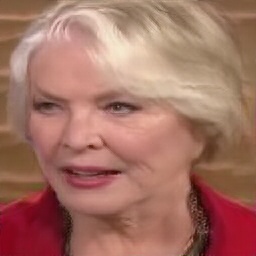}}
\\
\vspace{-0.95cm}
\subfloat[]{\includegraphics[height=0.14\linewidth, width=0.14\linewidth]{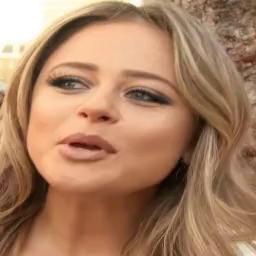}}
\subfloat[]{\includegraphics[height=0.14\linewidth, width=0.14\linewidth]{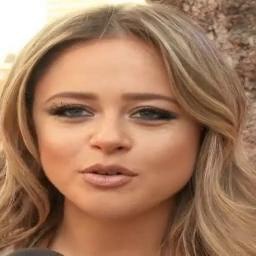}}\hspace{0cm}
\subfloat[]{\includegraphics[height=0.14\linewidth, width=0.14\linewidth]{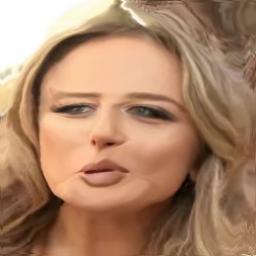}}
\subfloat[]{\includegraphics[height=0.14\linewidth, width=0.14\linewidth]{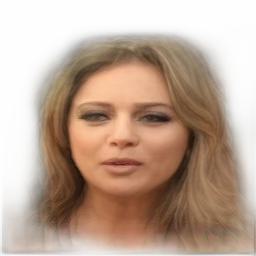}}
\subfloat[]{\includegraphics[height=0.14\linewidth, width=0.14\linewidth]{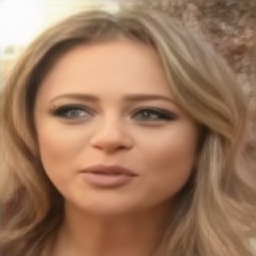}}
\subfloat[]{\includegraphics[height=0.14\linewidth, width=0.14\linewidth]{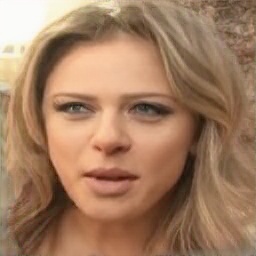}}
\subfloat[]{\includegraphics[height=0.14\linewidth, width=0.14\linewidth]{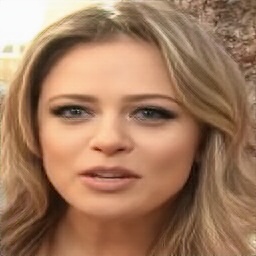}}
\\
\vspace{-0.95cm}
\subfloat[\tiny{Source}]{\includegraphics[height=0.14\linewidth, width=0.14\linewidth]{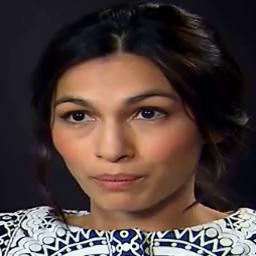}}
\subfloat[\tiny{Target}]{\includegraphics[height=0.14\linewidth, width=0.14\linewidth]{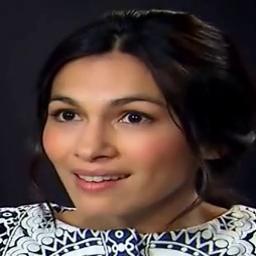}}\hspace{0cm}
\subfloat[\tiny{\textit{X2Face}\cite{wiles2018x2face}}]{\includegraphics[height=0.14\linewidth, width=0.14\linewidth]{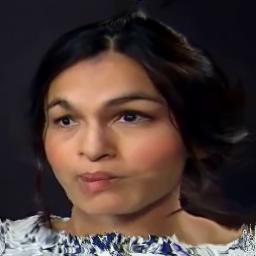}}
\subfloat[\tiny{\textit{Bi-layer}\cite{zakharov2020fast}}]{\includegraphics[height=0.14\linewidth, width=0.14\linewidth]{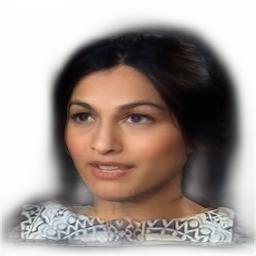}}
\subfloat[\tiny{\textit{FOMM}\cite{NEURIPS2019_31c0b36a}}]{\includegraphics[height=0.14\linewidth, width=0.14\linewidth]{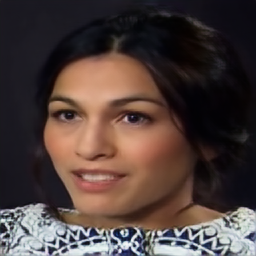}}
\subfloat[\tiny{\textit{PIRenderer}\cite{ren2021pirenderer}}]{\includegraphics[height=0.14\linewidth, width=0.14\linewidth]{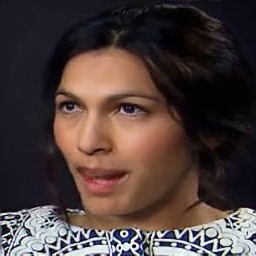}}
\subfloat[\tiny{Ours}]{\includegraphics[height=0.14\linewidth, width=0.14\linewidth]{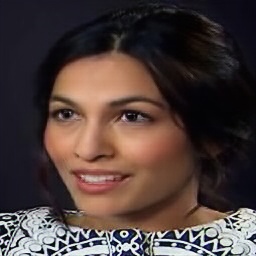}}
\caption{\textbf{Qualitative results of Same-identity reenactment with state-of-the-art methods.} The first column is the source image and the second column is the target image which refer to the target motion. And from the third to the last column are the result images of \textit{X2face}~\cite{wiles2018x2face}, \textit{Bi-layer}~\cite{zakharov2020fast}, \textit{FOMM}~\cite{NEURIPS2019_31c0b36a}, \textit{PIRenderer}~\cite{ren2021pirenderer}, and Ours, respectively.}
\label{fig:qulitative}
\end{figure*}
\section{Experiments}
\subsection{Dataset.}
We trained and evaluated our method on VoxCeleb dataset~\cite{nagrani2017voxceleb}. This dataset consists of videos in which faces move naturally within a bounding box collected from YouTube. From it, we exploited total 18,911(18,408 for training and 503 for testing) talking-head videos. Each faces are cropped and resized following the method of ~\cite{NEURIPS2019_31c0b36a}, and its facial motions are captured from it.

\subsection{Evaluation Metrics.}
We used basic image reconstruction measurement for the same identity reenactment comparison.
And we measured using the methods used in the previous papers~\cite{doukas2021headgan,ren2021pirenderer} for cross-identity reenactment because the ground-truth target images are not existed.

\noindent\textbf{Same-identity reenactment.}
We employ the \textit{Learned Perceptual Image Patch Similarity}~\cite{zhang2018unreasonable} (LPIPS) which is used for the estimation of reconstruction error between the generated and the ground-truth images.
Furthermore, we measure the realisticity of the generated frames by \textit{Fr$\acute{\text{e}}$chet Inception Distance}~\cite{NIPS2017_8a1d6947} (FID).
And we used \textit{Average Expression Distance} (AED) and \textit{Average Pose Distance} (APD) in the same way as \textit{PIRenderer}~\cite{ren2021pirenderer} for the motion accuracy evaluation.
The motion accuracy is evaluated by calculating the distance of the average 3DMM expressions and poses between the final output images and the target images as AED and APD respectively. 

\noindent\textbf{Cross-identity reenactment.}
We used three metrics to evaluate cross identity reenactment generated images, which do not have ground truth images.
First, \textit{Fr$\acute{\text{e}}$chet Inception Distance} (FID)~\cite{NIPS2017_8a1d6947} was used in the same way as the same identity reconstruction measure to calculate the data distribution between the target images and the final output images.
In addition, we used \textit{Cosine Similarity} (CSIM) to evaluate identity preservation.
When measuring CSIM in cross identity, we compared source images and final output images because there are no ground truth images that match the generated image.
In the same way as \textit{HeadGAN}~\cite{doukas2021headgan}, \textit{ArcFace}~\cite{deng2019arcface} was used as an identity recognition network. After extracting the features through the identity recognition network, we calculated the cosine similarity between the features of the source images and the final output images.
Finally, \textit{Average Pose Distance} (AED) also used to evaluate how well the expression was delivered in cross identity.
In cross identity, \textit{Average Pose Distance} (APD) was excluded because the starting pose values of the source images and the target images were different, so the corresponding comparison could not be made.

\subsection{Comparison with baselines}

\begin{figure*}[t!]
\captionsetup[subfigure]{labelformat=empty}
\centering
\subfloat[]{\includegraphics[height=0.14\linewidth, width=0.14\linewidth]{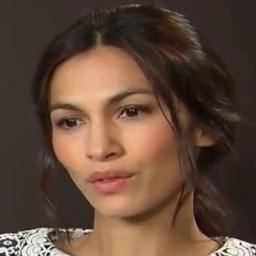}}
\subfloat[]{\includegraphics[height=0.14\linewidth, width=0.14\linewidth]{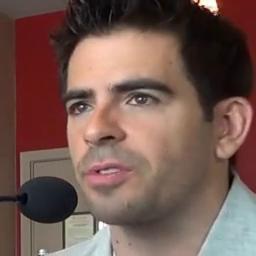}}\hspace{0cm}
\subfloat[]{\includegraphics[height=0.14\linewidth, width=0.14\linewidth]{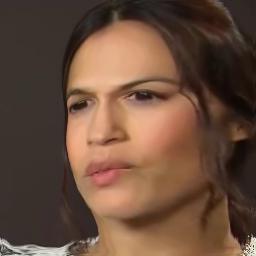}}
\subfloat[]{\includegraphics[height=0.14\linewidth, width=0.14\linewidth]{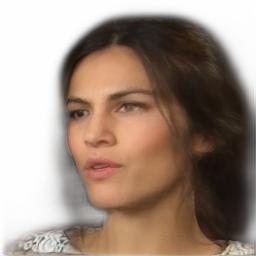}}
\subfloat[]{\includegraphics[height=0.14\linewidth, width=0.14\linewidth]{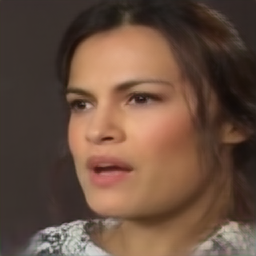}}
\subfloat[]{\includegraphics[height=0.14\linewidth, width=0.14\linewidth]{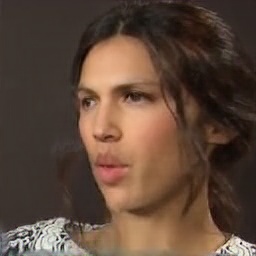}}
\subfloat[]{\includegraphics[height=0.14\linewidth, width=0.14\linewidth]{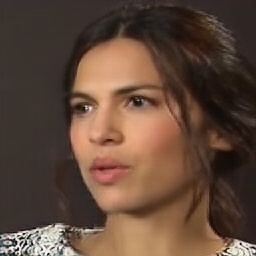}}
\\
\vspace{-0.95cm}
\subfloat[]{\includegraphics[height=0.14\linewidth, width=0.14\linewidth]{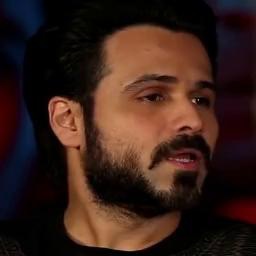}}
\subfloat[]{\includegraphics[height=0.14\linewidth, width=0.14\linewidth]{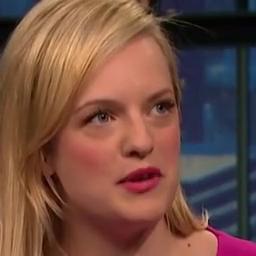}}\hspace{0cm}
\subfloat[]{\includegraphics[height=0.14\linewidth, width=0.14\linewidth]{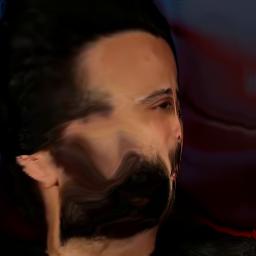}}
\subfloat[]{\includegraphics[height=0.14\linewidth, width=0.14\linewidth]{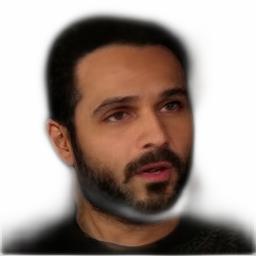}}
\subfloat[]{\includegraphics[height=0.14\linewidth, width=0.14\linewidth]{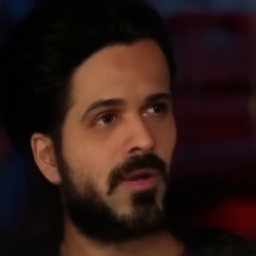}}
\subfloat[]{\includegraphics[height=0.14\linewidth, width=0.14\linewidth]{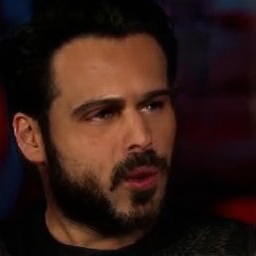}}
\subfloat[]{\includegraphics[height=0.14\linewidth, width=0.14\linewidth]{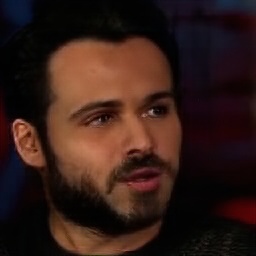}}
\\
\vspace{-0.95cm}
\subfloat[]{\includegraphics[height=0.14\linewidth, width=0.14\linewidth]{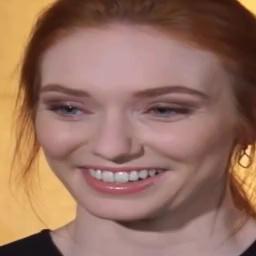}}
\subfloat[]{\includegraphics[height=0.14\linewidth, width=0.14\linewidth]{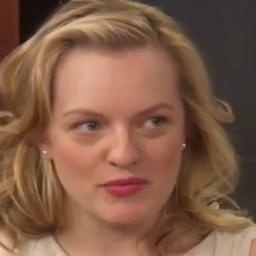}}\hspace{0cm}
\subfloat[]{\includegraphics[height=0.14\linewidth, width=0.14\linewidth]{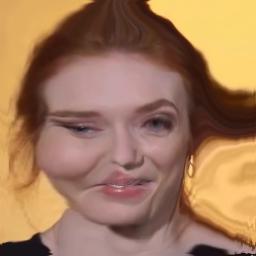}}
\subfloat[]{\includegraphics[height=0.14\linewidth, width=0.14\linewidth]{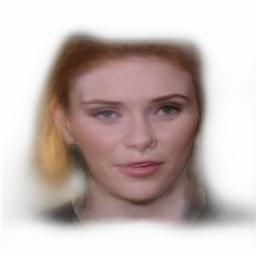}}
\subfloat[]{\includegraphics[height=0.14\linewidth, width=0.14\linewidth]{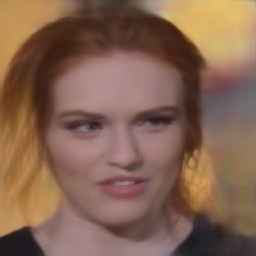}}
\subfloat[]{\includegraphics[height=0.14\linewidth, width=0.14\linewidth]{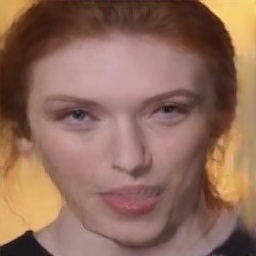}}
\subfloat[]{\includegraphics[height=0.14\linewidth, width=0.14\linewidth]{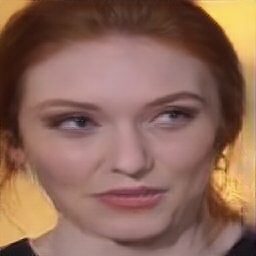}}
\\
\vspace{-0.95cm}
\subfloat[]{\includegraphics[height=0.14\linewidth, width=0.14\linewidth]{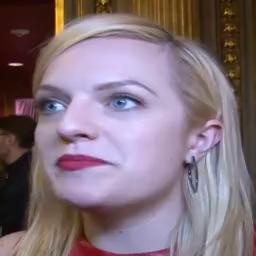}}
\subfloat[]{\includegraphics[height=0.14\linewidth, width=0.14\linewidth]{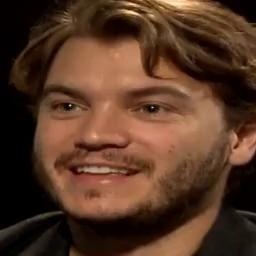}}\hspace{0cm}
\subfloat[]{\includegraphics[height=0.14\linewidth, width=0.14\linewidth]{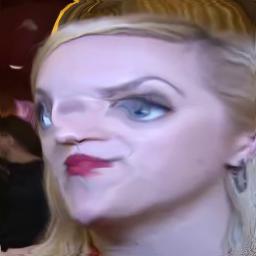}}
\subfloat[]{\includegraphics[height=0.14\linewidth, width=0.14\linewidth]{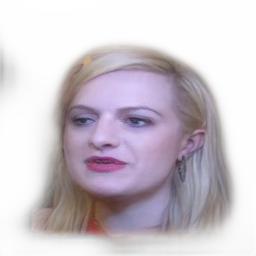}}
\subfloat[]{\includegraphics[height=0.14\linewidth, width=0.14\linewidth]{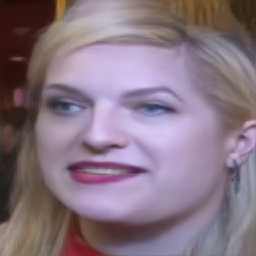}}
\subfloat[]{\includegraphics[height=0.14\linewidth, width=0.14\linewidth]{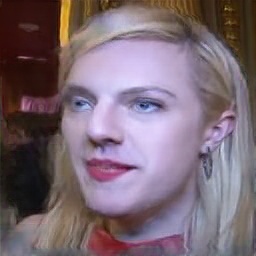}}
\subfloat[]{\includegraphics[height=0.14\linewidth, width=0.14\linewidth]{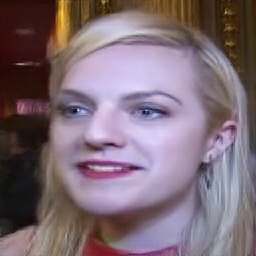}}
\\
\vspace{-0.95cm}
\subfloat[\tiny{Source}]{\includegraphics[height=0.14\linewidth, width=0.14\linewidth]{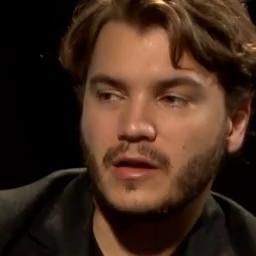}}
\subfloat[\tiny{Target}]{\includegraphics[height=0.14\linewidth, width=0.14\linewidth]{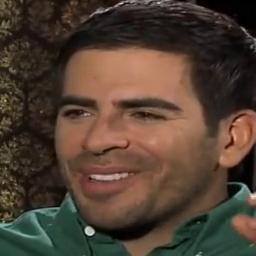}}\hspace{0cm}
\subfloat[\tiny{\textit{X2Face}\cite{wiles2018x2face}}]{\includegraphics[height=0.14\linewidth, width=0.14\linewidth]{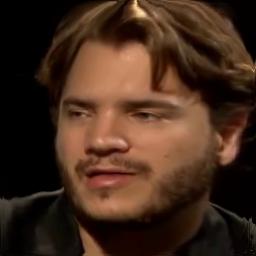}}
\subfloat[\tiny{\textit{Bi-layer}\cite{zakharov2020fast}}]{\includegraphics[height=0.14\linewidth, width=0.14\linewidth]{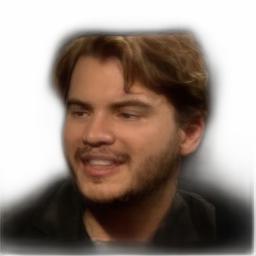}}
\subfloat[\tiny{\textit{FOMM}\cite{NEURIPS2019_31c0b36a}}]{\includegraphics[height=0.14\linewidth, width=0.14\linewidth]{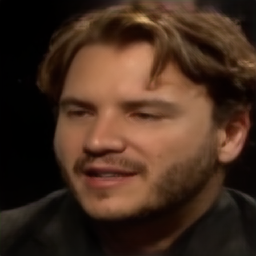}}
\subfloat[\tiny{\textit{PIRenderer}\cite{ren2021pirenderer}}]{\includegraphics[height=0.14\linewidth, width=0.14\linewidth]{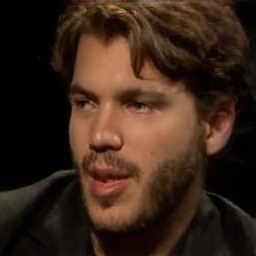}}
\subfloat[\tiny{Ours}]{\includegraphics[height=0.14\linewidth, width=0.14\linewidth]{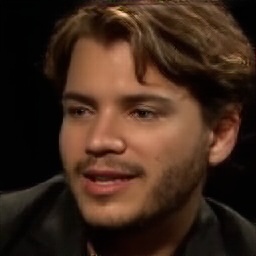}}
\caption{\textbf{Qualitative results of Cross-identity reenactment with state-of-the-art methods.} The first column is the source image and the second column is the target image to refer to the target motion. And from the third to the last column are the result images of \textit{X2face}~\cite{wiles2018x2face}, \textit{Bi-layer}~\cite{zakharov2020fast}, \textit{FOMM}~\cite{NEURIPS2019_31c0b36a}, \textit{PIRenderer}~\cite{ren2021pirenderer} and Ours, respectively.}
\label{fig:qulitative_cross}
\end{figure*}

We qualitatively and quantitatively compared our model with \textit{X2Face}~\cite{wiles2018x2face}, \textit{Bi-layer}~\cite{zakharov2020fast}, \textit{FOMM}~\cite{NEURIPS2019_31c0b36a}, and \textit{PIRenderer}~\cite{ren2021pirenderer}. 
Qualitative comparisons were made for the same identity reenactment and the cross identity reenactment, respectively, and qualitative results are shown in Fig.~\ref{fig:qulitative} and Fig.~\ref{fig:qulitative_cross}. The quantitative evaluation results can be confirmed through Table \ref{table1}. 
We used pre-trained models provided by authors for~\cite{wiles2018x2face},~\cite{NEURIPS2019_31c0b36a}, and~\cite{ren2021pirenderer}, and these models were trained using VoxCeleb dataset. 
And for the~\cite{zakharov2020fast}, the pre-trained model provided was also used, which was trained with the VoxCeleb2~\cite{chung2018voxceleb2}, which is the larger dataset.

\noindent\textbf{Same-identity reenactment.}
Same identity reenactment is the task of performing a reconstruction with the first frame of the video as the source image and the remaining frames as the target images.
In the same identity reenactment, we compared 503 test videos frame by frame.
A total of around 146K images were compared with each method.
Since the \textit{Bi-layer}~\cite{zakharov2020fast} does not generate a background part of the image, we measured its performance with an image excluding the background image.
The results of our method outperform the other methods in  same-identity reenactment, as shown in the quantitative results Table~\ref{table1} and qualitative results Fig.~\ref{fig:qulitative}.

\noindent\textbf{Cross-identity Reenactment.}
In the cross identity reenactment, we randomly selected 20 pairs of different identities videos out of 503 test videos.
Then, by setting the 0-th frame of the source video as the source image and the target video with different identities as the target motion, images for each method were generated.
For each method, 20 videos were created.
And we compared the selected video by frame.
A total of around 4K images were compared with each method.
Since the \textit{Bi-layer}~\cite{zakharov2020fast} does not generate the background part of the image, the cross-identity reenactment was also measured with the image excluding the background image in the quantitative evaluation. 
Table~\ref{table1} and Fig.~\ref{fig:qulitative_cross} show quantitative and qualitative evaluations. 
In cross-identity reenactment, the results of our method also outperform other methods, as can be seen in the quantitative results Table~\ref{table1} and qualitative results Fig.~\ref{fig:qulitative_cross}.

\begin{table}[t]
\centering
\caption{Quantitative results for the VoxCeleb test set~\cite{nagrani2017voxceleb} on same-identity reenactment and cross-identity reenactment tasks with state-of-the-art methods.}
\vspace{0.3cm}
\label{table1}
\resizebox{\columnwidth}{!}{
\begin{tabular}{l|cccc|cccc}
\cline{1-9} 
&\multicolumn{4}{c}{Same-identity Reenactment} & \multicolumn{4}{|c}{Cross-identity Reenactment}\\
\cline{2-9}
&FID$\downarrow$&LPIPS$\downarrow$&AED$\downarrow$&APD$\downarrow$&FID$\downarrow$&CSIM$\uparrow$&AED$\downarrow$&APD$\downarrow$\\
\hline
\noalign{\smallskip}
\hline
\textit{X2face}~\cite{wiles2018x2face} &32.20&0.240&0.168&0.115&65.04&0.450&0.250&0.141\\
\textit{Bi-layer}~\cite{zakharov2020fast} &44.92&0.166&0.113&0.054&90.43&0.463&0.183&0.070\\
\textit{FOMM}~\cite{ren2021pirenderer} &14.01&0.127&0.203&0.076&59.77&0.570&0.188&0.076\\
\textit{PIRenderer}~\cite{ren2021pirenderer} &22.55&0.191&0.136&0.074&59.58&0.524&0.210&0.096\\
Ours &\bf{10.36}&\bf{0.126}&\bf{0.082}&\bf{0.045}&\bf{45.26}&\bf{0.582}&\bf{0.156}&\bf{0.069}\\
\hline
\end{tabular}
}
\end{table}

\subsubsection{Audio-driven face reenactment.} 
\begin{figure*}[t!]
\centering
\includegraphics[width=1\linewidth]{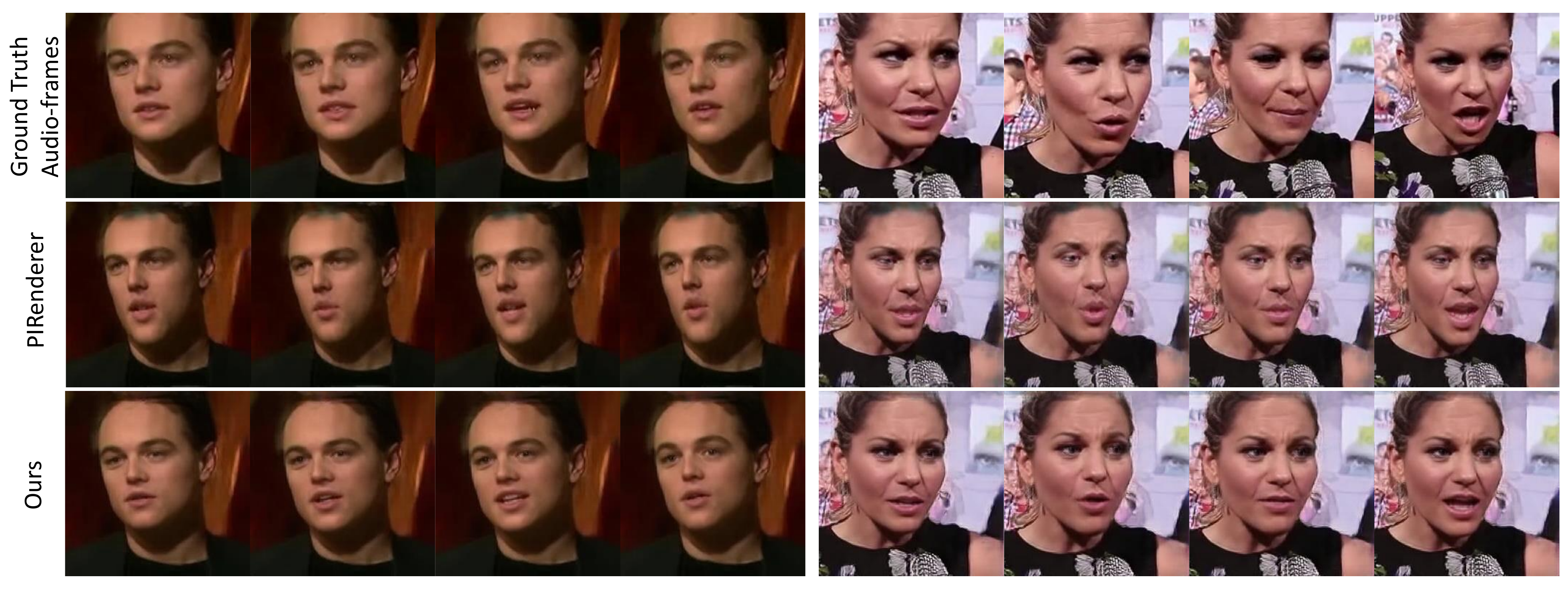}
\caption{\textbf{Examples results from Ours and competitive baseline 
PIRenderer~\cite{ren2021pirenderer} on VoxCeleb2,} which is not used in training on both methods. Our method shows better qualitative performance on audio-based face reconstruction tasks with the new dataset.}
\label{fig:fig5}
\end{figure*}

In this section, we performed an audio-driven face reenactment task on the baseline model \textit{PIRenderer}~\cite{ren2021pirenderer} and ours to qualitatively compare the results.
The audio-driven face reenactment was performed using VoxCeleb2~\cite{chung2018voxceleb2} dataset which was not used during the training process in both methods.
We set the 0-th frame in the video as the source image and the audio sources of the remaining frames as the target motion. As we used 3DMM as target motion in the same way as~\cite{ren2021pirenderer}, we generated images under the same conditions by extracting 3DMM parameters from the audio source and taking 3DMM parameters as target motion in each method. The qualitative results are shown in Fig.~\ref{fig:fig5}.  It shows that the result of our method more closely matches the target motion, resulting in a sharp image. This allowed us to test the expandability of the face reenactment task in conditions of intuitive control, using semantically independent parameters as target motion.

\section{Conclusions}
We introduced IFaceUV, a model that can intuitively generate real-world face images by modifying 3DMM parameters such as facial emotions, head posture, and translation. State-of-the-art face reenactment approaches struggle to maintain a facial identity while taking subject-agnostic target motion. Our method serves this problem by employing an approach that combines the advantages of an explicit 3D face model with the advantages of a 2D spatial transformation to generate realistic facial images that target specific motions. Our model is somewhat affected by performance of the 3D face reconstruction method. Nonetheless, when compared to state-of-the-art approaches, our model outperformed them in terms of visual quality. Our performance is expected to improve if the model of 3D face reconstruction is enhanced. In future work, we should investigate the possibility of extending our method to the sequential domain.

\clearpage
\bibliographystyle{splncs04}
\bibliography{egbib}      
\end{document}